%% file: main.tex
\documentclass[runningheads]{llncs}

\usepackage{eccv}
\usepackage{eccvabbrv}

\usepackage{graphicx}
\usepackage{booktabs}

\usepackage[accsupp]{axessibility}

\usepackage[pdfborder={0 0 0}]{hyperref}

\usepackage{orcidlink}

\usepackage{bm}
\usepackage{siunitx} 
\usepackage{multirow}
\usepackage{multicol} 
\usepackage{dsfont} 
\usepackage{graphicx} 
\usepackage[normalem]{ulem} 

\renewcommand{\paragraph}[1]{\textbf{#1}.}
\newcommand{\mbf}[1]{\mathbf{#1}}
\newcommand{\tbf}[1]{\textbf{#1}}

\usepackage[table]{xcolor}
\definecolor{bestcolor}{RGB}{255,179,179}     
\definecolor{secondcolor}{RGB}{255,204,153}   
\definecolor{thirdcolor}{RGB}{255,255,153}    
\newcommand{\best}[1]{\cellcolor{bestcolor}\textbf{#1}}
\newcommand{\second}[1]{\cellcolor{secondcolor}#1}
\newcommand{\third}[1]{\cellcolor{thirdcolor}#1}

\usepackage{titletoc} 

\begin{document}

\title{FLM-Occ: Feed-forward Likelihood Maximization \\for
	Efficient Indoor Occupancy Prediction}

\titlerunning{FLM for Indoor Occ}

\author{
	Guangcheng Chen\inst{1,3}\orcidlink{0000-0001-8191-3327} \and
	Lihuang Fang\inst{1,3}\orcidlink{0009-0006-6952-4891} \and
	Huaqi Tao\inst{1,3}\orcidlink{0000-0002-6839-6596} \and
	Yicheng He\inst{1,3}\orcidlink{0000-0003-3213-9907} \and \\
	Li He\thanks{Li He and Hong Zhang are corresponding authors.}\inst{1,2}\orcidlink{0000-0003-0261-4068} \and
	Hong Zhang\textsuperscript{\ensuremath{\star}}\inst{1,3}\orcidlink{0000-0002-1677-6132}
}

\authorrunning{G.~Chen, L.~Huang \etal.}

\institute{
	Southern University of Science and Technology \and
	Guangdong Laboratory of Artificial Intelligence and Digital Economy (Shenzhen) \and
	Shenzhen Key Laboratory of Robotics and Computer Vision\\
	\url{https://gcchen97.github.io/flm-occ/}
}

\maketitle

\input{sec/0_abstract}    
\input{sec/1_intro}

\input{sec/2_review}
\input{sec/3_method}

\input{sec/4_exp}

\input{sec/5_conclusion}
\input{sec/acknowl}

\bibliographystyle{splncs04}
\bibliography{main}

\clearpage
\appendix

\input{sec/X_suppl}

\input{sec/X_kernel_volume}

\end{document}

%% file: sec/0_abstract.tex
\begin{abstract}
Recent indoor occupancy prediction methods adopt Gaussian primitives as a sparse 3D representation for computational efficiency.
However, their training relies on voxel classification, which imposes only local constraints and lacks global supervision on the distribution of the primitives.
Therefore, they inevitably predict spurious primitives in empty regions, undermining both representational and computational efficiency.
To address this, we propose \underline{F}eed-forward \underline{L}ikelihood \underline{M}aximization (FLM), a novel framework that reformulates occupancy prediction as voxel distribution estimation.
In FLM, a network is trained to predict a mixture model that maximizes the likelihood over ground-truth occupied voxels in a feed-forward manner.
To enable end-to-end training of networks and voxelization of a standard mixture model, we define mixture weights as normalized primitive volumes to implicitly enforce simplex constraints and derive novel voxelization formulas.
Based on FLM, our FLM-Occ, a novel method that is capable of relocating randomly initialized primitives over long distances to model a scene.
On Occ-ScanNet, FLM-Occ achieves superior accuracy using only 32 superquadrics, $2.7\%$ of the prior SoTA, while running $3.7\times$ faster.

\end{abstract}

%% file: sec/1_intro.tex
\section{Introduction}
\label{sec:intro}

Monocular occupancy prediction aims to reconstruct a geometric and semantic 3D layout from a single RGB image in the form of a voxel grid~\cite{monoscene2022cvpr,ndcscene2023iccv}.
It has been widely-studied in autonomous driving~\cite{openocc2023iccv,sparseocc2024cvpr,zuo2025gaussianworld} and embodied AI~\cite{sscnet2017cvpr,ndcscene2023iccv,embodiedocc2025iccv,discene2025ral}.
Yet dense occupancy prediction suffers from cubic computational complexity, making it prohibitively expensive for real-time applications.

\begin{figure}[t]
	\centering
	\includegraphics[width=122mm]{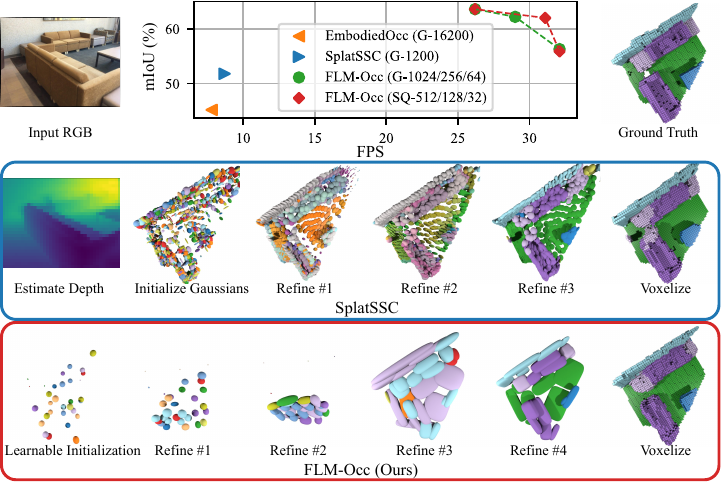}
	\caption{
		\tbf{Comparison (top-left) on Occ-ScanNet~\cite{iso2024eccv}.} The numbers following each method name indicate primitive count. Our FLM-Occ achieves significantly better accuracy–efficiency trade-offs than EmbodiedOcc~\cite{embodiedocc2025iccv} and SplatSSC~\cite{splatssc2025}.
		The second and third rows show the prediction process of SplatSSC and FLM-Occ.
	}
	\label{fig:intro}
\end{figure}

\begin{figure}[t]
	\centering
	\includegraphics[width=122mm]{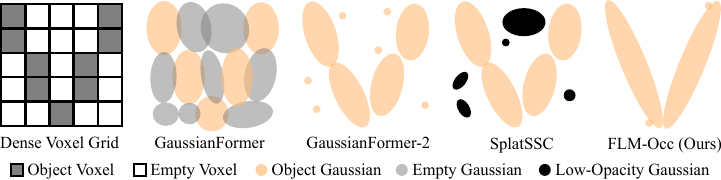}
	\caption{
		\tbf{Four Gaussian-based methods for occupancy prediction.}
		GaussianFormer~\cite{gaussianformer2024eccv}, GaussianFormer-2~\cite{gaussianformer22025cvpr}, and SplatSSC~\cite{splatssc2025} leave redundant primitives in empty regions because they are incapable of substantially relocating the Gaussians, undermining their representational efficiency.
		With MLE, FLM-Occ learns to relocate Gaussians toward occupied regions.
	}
	\label{fig:4mm}
\end{figure}

Recent methods have explored various 3D representations for efficiency.
Among them, mixture model-based methods~\cite{gaussianformer2024eccv,embodiedocc2025iccv,gaussianformer22025cvpr,quadricformer2025,splatssc2025,discene2025ral} have drawn increasing attention.
They require far fewer geometric primitives than sparse voxel-based methods~\cite{openocc2023iccv,voxformer2023cvpr,sparseocc2024cvpr,octmae2024eccv,octreeocc2024nips,voxelmamba2024nips,occmamba2025cvpr} because each primitive can represent multiple voxels.
They also provide continuous spatial representation as plane-based methods~\cite{tpvformer2023cvpr,flashocc2023,fastocc2024icra,sliceocc2025icra}, allowing self-supervision via image rendering~\cite{selfocc2024cvpr}.
During inference, they initialize a fixed number of primitives and refine them to model the semantic distribution of the scene in the input image.
However, as illustrated in \cref{fig:4mm}, they suffer from a critical limitation: spurious primitives persist in empty regions, wasting representation capacity and increasing computation.

In this paper, we identify the root cause of this problem as the training objective---voxel classification.
When these methods voxelize the predicted primitives, a voxel-wise cross-entropy loss is used for training.
However, this loss cannot optimize the global distribution of the primitives and fails to induce distant primitive relocations.
Our gradient analysis shows that this is because the training gradients of this loss with respect to primitive positions vanish when the primitives are far from ground-truth occupied voxels.
Consequently, their networks are only able to make local adjustments of the primitives.

To resolve the above problem, we propose Feed-forward Likelihood Maximization (\textbf{FLM}), a novel framework that reformulates occupancy prediction as voxel distribution estimation, parameterized by a mixture model.
This framework is based on maximum likelihood estimation (MLE)~\cite{mle2003}, which is a widely used method for estimating the parameters that best explain the observed data.
Specifically, our FLM objective maximizes the joint likelihood of all ground truth occupied voxels, and the trained network learns to predict a mixture model that maximizes the likelihood in a feed-forward manner.

As it turns out, enabling end-to-end training in FLM is non-trivial since simplex constraints are required for mixture weights and a new voxelization method is necessary for a standard mixture model.
To tackle this issue, we define \emph{the weights of the primitives as normalized primitive volumes}.
This implicitly enforces the simplex constraints, enabling end-to-end training.
In addition, it aligns each mixture weight with the spatial extent of the primitive, preventing degenerate cases where excessively large primitives receive vanishing weights and thereby preserving representational efficiency.

Finally, we present \textbf{FLM-Occ}, a lightweight network for monocular occupancy prediction.
Unlike prior work that relies on depth maps to place the primitives, FLM-Occ directly refines them with the guidance of image features to match scene structures.
FLM-Occ is able to substantially relocate randomly initialized primitives to align with objects.
As shown in \cref{fig:intro},
FLM-Occ achieves compact representation, efficient inference, and superior performance on the Occ-ScanNet dataset~\cite{iso2024eccv}.

%% file: sec/2_review.tex
\section{Related Work}
\label{sec:ch2}

\noindent\textbf{Mixture Model-based Occupancy Prediction}.
Mixture model-based methods train neural networks to refine a set of primitives, such as Gaussians~\cite{gaussianformer2024eccv} or superquadrics~\cite{quadricformer2025}, to model a scene.
The pioneering work GaussianFormer~\cite{gaussianformer2024eccv} uniformly distributes Gaussians in 3D space and then deforms and classifies them to model the semantic distribution.
The occupancy grid for calculating training loss is then extracted from the Gaussians.
GaussianFormer~\cite{gaussianformer2024eccv} does not take advantage of the sparsity of 3D scenes for optimizing its efficiency, as it wastefully models empty regions with Gaussians.
Later work exploits scene sparsity by using object-centric representations with fewer primitives.
GaussianFormer-2~\cite{gaussianformer22025cvpr} predicts the distributions of objects along camera rays to initialize Gaussians accordingly.
SplatSSC~\cite{splatssc2025} uses a depth estimation model to predict depth maps and places initial Gaussians at the estimated depths.
Nevertheless, their primitives are adjusted locally, and thus some of them persist in empty regions.
We address this problem by leveraging MLE for training, enabling distant primitive relocations and thus eliminating the need for initializing primitives near objects.

\paragraph{Training Objectives of Occupancy Prediction}
Occupancy prediction has long been formulated as a voxel classification problem, regardless of the underlying 3D representation.
OPUS~\cite{opus2024nips} alternatively adopts a point regression paradigm, yet it still relies on 
voxel-wise regression and operations that remain computationally expensive.
Other methods~\cite{bts2023cvpr,selfocc2024cvpr,gaussianocc2025iccv,voxelsplat2025cvpr,scenedino2025iccv} adopt image reconstruction losses~\cite{nerf2020eccv,3dgs2023sig,monogs2024cvpr} for self-supervision, but their performance remains limited.
In contrast, we reformulate occupancy prediction as voxel distribution estimation and train a network via MLE,
enabling global supervision on the mixture models and encouraging large relocations of the primitives.
This formulation allows us to fully exploit the representation capacity of a mixture model and thereby minimizes the number of primitives and accelerates inference.

\paragraph{Feed-forward 3D Gaussian Splatting (F3DGS)}
Since the pioneering work 3D Gaussian Splatting (3DGS)~\cite{3dgs2023sig}, mixture models have emerged as a transformative alternative representation in 3D reconstruction.
Rather than estimating the parameters of a mixture model through iterative optimization, F3DGS methods~\cite{splatter2024cvpr,pixelsplat2024cvpr,mvsplat2024eccv,gslrm2024eccv,noposplat2025iccv} train networks to estimate the parameters in a feed-forward manner.
However, these methods are unable to substantially relocate the Gaussians.
In 3DGS~\cite{3dgs2023sig}, Gaussians are indirectly relocated through the non-differentiable densification and pruning~\cite{3dgs2023sig,pixelsplat2024cvpr} process.
As for F3DGS methods, they avoid this problem by representing scenes with pixel-aligned Gaussian maps~\cite{splatter2024cvpr} that align well with their pixel-wise training losses.
In our work, we instead leverage MLE to learn global relocation of unordered primitives.

%% file: sec/3_method.tex
\section{Methodology}
We present our FLM and its application in monocular occupancy prediction for indoor scenes.
We begin by reviewing the formulations of prior mixture model-based methods and their training objective in Section~\ref{ch3-prelim}.  
We then introduce FLM in Sections~\ref{ch3-FLM}, followed by the architecture of FLM-Occ in Section~\ref{ch3-network}.

\subsection{Preliminaries}
\label{ch3-prelim}

Mixture model-based methods~\cite{gaussianformer2024eccv,embodiedocc2025iccv,gaussianformer22025cvpr,quadricformer2025,splatssc2025} refine a fixed number of primitives, denoted by $\{\mathbf{g}_j\}_{j=1}^M$, to model 3D scenes according to the input images.
The primitive positions can be initialized either by uniform sampling in 3D space~\cite{gaussianformer2024eccv} or by back-projecting samples from a depth map~\cite{gaussianformer22025cvpr, splatssc2025}.
The number of primitives is fixed during training and inference.

These methods can be categorized into two types, depending on whether they use object-centric representations.
The pioneering work GaussianFormer~\cite{gaussianformer2024eccv} uses Gaussians to model scenes.
Each Gaussian $\mathbf{g}_j$ consists of a position $\bm{\mu}_j\in\mathbb{R}^3$, scale $\mathbf{s}_j\in\mathbb{R}^3$, a rotation $\mathbf{R}_j\in\mathbb{R}^{3\times 3}$, an opacity $a_j$ and  semantic logits $\mathbf{f}_j\in\mathbb{R}^{n}$.
The semantic logits $\mathbf{c}$ of voxel $\mathbf{x}$ are aggregated from neighboring Gaussians:
\begin{gather}
	\label{eq:gformer}
	\mathbf{c}=\sum_{j\in\mathcal{N}} a_j\mathcal{K}(\mathbf{x}|\mathbf{g}_j)\mathbf{f}_j,
\\
	\label{eq:N}
	\mathcal{N} = \big\{ j \in \{1, 2, \dots, M\} \mid \|\mathbf{x} - \bm{\mu}_j\|_{\infty} < 3 \|\mathbf{s}_j\|_{\infty} \big\},
\end{gather}
where $\mathcal{K}$ is the primitive function and $\mathcal{N}$ is the index set of neighboring Gaussians.
$\mathcal{N}$ is used for reducing computation.
This representation is more efficient than voxels because each Gaussian can present several voxels, but it still models empty regions with Gaussians.

\input{tables/ch3-comparison}

GaussianFormer-2~\cite{gaussianformer22025cvpr} improves its predecessor by using an object-centric representation that separately models geometry and semantics,
aiming to restrict Gaussians in occupied regions for efficiency.
The probability of a voxel being occupied $p$ and the semantics $\mathbf{c}$ are calculate as
\begin{equation}
	\label{eq:gformer2}
	p=1-\prod_{j\in\mathcal{N}}\big(1-\mathcal{K}(\mathbf{x}|\mathbf{g}_j)\big),\,
	\mathbf{c}=
		\frac{
			\sum_{j\in\mathcal{N}}\frac{a_j}{v_j}\mathcal{K}(\mathbf{x}|\mathbf{g}_j)\mathbf{f}'_j
		}{
			\sum_{j\in\mathcal{N}}\frac{a_j}{v_j}\mathcal{K}(\mathbf{x}|\mathbf{g}_j)
		},
\end{equation}
where $a_j\in[0,1]$, is opacity, $\mathbf{f}'_j$ is the softmax-normalized $\mathbf{f}_j$, $v_j$ is the volume of primitive $\mathbf{g}_j$ calculated as:
\begin{equation}
	v_j=\int_{\mathbb{R}^3} \mathcal{K}(\mathbf{x}|\mathbf{g}_j)d\mathbf{x}.
\end{equation}
Later work improves \cref{eq:gformer2} by utilizing superquadrics~\cite{quadricformer2025} or devising more effective primitive weighting~\cite{splatssc2025}.
\cref{tb:mm} summarizes the different methods.

Regardless of the voxelization formulas, both types of methods calculate the same voxel classification loss $\mathcal{L}_{vc}$ 
for ``empty'' and semantic classes with the cross-entropy function $f_{ce}$ on the output ${H\times W\times D}$ voxel grid :
\begin{equation}
	\label{eq:voxelloss}
	\mathcal{L}_{vc}=\frac{1}{HWD}\sum_{i}^{HWD}f_{ce}(\mathbf{c}'_i,\mathbf{c}_i^{(gt)}).
\end{equation}
The only difference lies in the definition of $\mathbf{c}'_i$: type 1 uses $\mathbf{c}'_i=\mathbf{c}_i$ while type 2 uses $\mathbf{c}'_i=[1-p_i;p_i\cdot\mathbf{c}_i]$.

\begin{figure}[t]
	\centering
	\includegraphics[width=97.6mm]{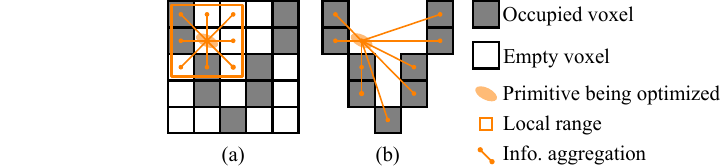}
	\caption{
	\tbf{
	Information graphs of the training losses for a single primitive via voxel classification (a) and MLE (b).
	}
	}
	\label{fig:local_global}
\end{figure}

A critical limitation of $\mathcal{L}_{vc}$ is its inability to relocate primitives to distant locations.
This limitation can be shown by analyzing its gradient.
We consider the case of binary classification without loss of generality and derive the gradient of $\mathcal{L}_{vc}$ for a ground-truth occupied voxel $\mathbf{x}_i$ w.r.t. the position of $\mathbf{g}_j$:
\begin{equation}
\label{eq:gf2grad}
	\frac{\partial \mathcal{L}^{(i)}_{vc}}{\partial \bm{\mu}_j}
	=
	-\frac{\prod_{r\ne j,r\in\mathcal{N}}\big(1-\mathcal{K}(\mathbf{x}_i|\mathbf{g}_r)\big)}
	{1-\prod_{r\in\mathcal{N}}\big(1-\mathcal{K}(\mathbf{x}_i|\mathbf{g}_r)\big)}
	\cdot
	\frac{\partial \mathcal{K}(\mathbf{x}_i|\mathbf{g}_j)}{\partial \bm\mu_j},
\end{equation}
with $\mathcal{K}_r=\mathcal{K}(\mathbf{x}_i|\mathbf{g}_r)$ and $\mathcal{K}_j=\mathcal{K}(\mathbf{x}_i|\mathbf{g}_j)$.
This suggest that the gradient $\partial \mathcal{L}^{(i)}_{vc}/\partial \bm{\mu}_j$ vanishes when any $\mathcal{K}_r$ is close to $1$.
In other words, if an occupied voxel is already explained by neighboring primitives, as shown in \cref{fig:local_global}, this voxel cannot impact other primitives, leaving distant primitives stranded in empty regions with negligible gradients.

\subsection{Feed-forward Likelihood Maximization}
\label{ch3-FLM}
To overcome the limitation of voxel-wise classification loss, we revisit MLE, a widely used method for estimating probabilistic models.
The goal of MLE is to find the parameters that maximize the joint probability for the observed data.
This aligns well with the goal of occupancy prediction, i.e., to predict a mixture model that accurately fits the distribution of the ground-truth occupied voxels:
\begin{gather}
	\label{eq:MLE}
	\{\hat{\mathbf{g}}_j\}_{j=1}^M=
	\arg\max_{\{\mathbf{g}_j\}_{j=1}^M}
	\prod_{i=1}^P\sum_{j=1}^M 
	\frac{w_j}{v_j}
	\mathcal{K}(\mathbf{x}_i|\mathbf{g}_j)\\
	\label{eq:simplex}
	\text{subject to}\quad
	\sum_{j=1}^{M}w_j=1
	\ .
\end{gather}
where $w_j\in[0,1]$ is the mixture weight
and $\{\mathbf{x}_i\}^P_{i=1}$ is the positions of the occupied voxels.
$\mathcal{K}$ could be any unnormalized probabilistic kernel function.

\cref{eq:MLE} forms the basis of our Feed-forward Likelihood Maximization (FLM).
Within FLM, a neural network $\mathcal{F}_\theta$ is trained to refine an initial mixture model $\{\mathbf{g}_j^{(0)}\}_{j=1}^M$, with the guidance of image features $\mathcal{F}_{enc}(\mathcal{I})$, such that the resulting model maximizes the scene likelihood:
\begin{equation}
	\{\hat{\mathbf{g}}_j\}_{j=1}^M=\mathcal{F}_\theta\Big(\{\mathbf{g}_j^{(0)}\}_{j=1}^M,\,\mathcal{F}_{enc}(\mathcal{I})\ \Big)\ .
\end{equation}
However, the end-to-end training of $\mathcal{F}_\theta$ is non-trivial since directly utilizing \cref{eq:MLE} for training raises two issues: how to (1) handle the simplex constraint \cref{eq:simplex} and (2) voxelize a standard mixture model.
These issues do not arise in previous methods because they replace mixture weights with unconstrained opacities and use different voxelization formulas designed for classification loss.
To address them, we reparameterize mixture weights as normalized volumes:
\begin{equation}
	\label{eq:volume-weight}
	w_j=\frac{v_j}{\sum^M_{j=1}v_j}\ .
\end{equation}
This reparameterization implicitly enforces simplex constraints on the weights.
It also ensures that each primitive's weight is proportional to its spatial extent so that the total volume $\sum^M_{j=1}v_j$ directly reflects the number of occupied voxels.
Consequently, large primitives account for big structures, while smaller ones specialize in local details.
With this reparameterization, the joint likelihood in \cref{eq:MLE} can be rewritten as:
\begin{equation}
	\label{eq:VM}
	\Bigg(
	\prod_{i=1}^P\frac{1}{\sum^M_{j=1}v_j}
	\Bigg)
	\cdot
	\Bigg(
	\prod_{i=1}^P
	\sum_{j=1}^M
	\mathcal{K}(\mathbf{x}_i|\mathbf{g}_j)
	\Bigg)
	.
\end{equation}

\begin{figure}[t]
	\centering
	\includegraphics[width=81.3mm]{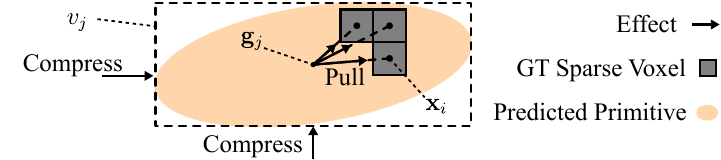}
	\caption{
		\textbf{2D illustrations of the roles of the two terms in the FLM loss.}
		The volume term compresses the primitives to match the voxel shape,
		and the density term pulls them toward voxel locations.
	}
	\label{fig:compress_pull}
\end{figure}

\paragraph{FLM Loss}
To construct a loss for numerically stable training, we take the negative logarithm of the likelihood in \cref{eq:VM} and average it over occupied voxels, resulting in the following FLM loss:
\begin{equation}
	\label{eq:flmloss}
	\mathcal{L}_{flm}=
	\underbrace{
		\log\sum_{j=1}^M v_j
	}_{\text{Volume sum}}
	-
	\underbrace{
		\frac{1}{P}
		\sum_{i=1}^P
		\log
		\sum_{j=1}^M
		\mathcal{K}(\mathbf{x}_i|\mathbf{g}_j)
	}_{\text{Density sum}}
	.
\end{equation}
Interestingly, 
minimizing the volume term of FLM loss will compress the total volume, thereby reducing densities in empty regions, and increasing the density term will pull the primitives toward occupied voxels.
In this sense, it explicitly drives all primitives toward objects and fully exploits their representational capacity.
Therefore, using FLM loss allows us to use significantly fewer primitives than using $\mathcal{L}_{vc}$ \big(\cref{eq:voxelloss}\big).
\cref{fig:compress_pull} illustrates the physical interpretation of the two terms of FLM loss.
To understand why $\mathcal{L}_{flm}$ is able to overcome the limitation of $\mathcal{L}_{vc}$, we derive the gradients of FLM loss of $\mathbf{x}_i$ w.r.t. the position of $\mathbf{g}_j$:
\begin{equation}
\label{eq:flmgrad}
	\frac{\partial\mathcal{L}^{(i)}_{flm}}{\partial\bm\mu_j}
	=
	-\frac{1}{\sum^{M}_{j=1}\mathcal{K}(\mathbf{x}_i|\mathbf{g}_j)}
	\cdot
	\frac{\partial \mathcal{K}(\mathbf{x}_i|\mathbf{g}_j)}{\partial \bm\mu_j}.
\end{equation}
Compared to $\partial\mathcal{L}^{(i)}_{vc}/\partial \bm\mu_j$ \big(\cref{eq:gf2grad}\big), $\partial\mathcal{L}^{(i)}_{flm}/\partial \bm\mu_j$ does not vanish when $\mathcal{K}(\mathbf{x}_i|\mathbf{g}_j)\to 1$, and therefore every $\mathbf{x}_i$ can provide relocation signals to all primitives.
So, $\mathcal{L}_{flm}$ can supervise the global distribution of the primitives and enable a network to generate primitive relocation when necessary.
Moreover, the computational cost of $\mathcal{L}_{flm}$ is also much less than that of $\mathcal{L}_{vc}$ because it only considers occupied regions, especially in Occ-ScanNet~\cite{iso2024eccv} where only 3\% of voxels are occupied.

We further unify the closed forms of $v_j$ for Gaussian and superquadric:
\begin{equation}
	v_j = 2^{\tfrac{3\epsilon_1}{2}}\;\epsilon_1^2\,\epsilon_2\;
	\frac{\Gamma\!\big(\tfrac{\epsilon_2}{2}\big)^2}{\Gamma(\epsilon_2)}\;
	\Gamma(\epsilon_1)\;\Gamma\!\big(\tfrac{\epsilon_1}{2}\big)s_x s_y s_z,
\end{equation}
where $(\epsilon_1,\epsilon_2)^\top$ and $(s_x,s_y,s_z)^\top$ are shape parameters and scales, respectively, and $\Gamma(z)=\int_0^\infty e^{-t}t^{z-1}dt$ is the gamma function~\cite{gamma}.
When $\epsilon_1=\epsilon_2=1$, $v_j=(2\pi)^{\frac{3}{2}}s_xs_ys_z$ is the volume of the Gaussian.
The derivation is provided in the supplementary material.

\paragraph{Voxelization}
The density term in \cref{eq:flmloss} naturally leads to effective and object-centric voxelization formulas for geometry and semantics:
\begin{equation}
	\label{eq:geosem}
	p=
	\sum_{j=1}^M\mathcal{K}(\mathbf{x}|\mathbf{g}_j),
	\,
	\mathbf{c}=
	\frac{\sum_{j=1}^{M}\mathcal{K}(\mathbf{x}|\mathbf{g}_j)\mathbf{f}_j}
	{\sum_{j=1}^{M}\mathcal{K}(\mathbf{x}|\mathbf{g}_j)}
	.
\end{equation}
Notably, $\mathbf{c}$ is the sum of the contributions that reflect how much each primitive contributes to the semantics of $\mathbf{x}$.

\paragraph{Regularization}
During voxelization, a point is classified as occupied if its density exceeds a predefined threshold.
Empirically, we find that most density values lie in $[0,1]$ and a global threshold of $0.5$ yields better results (see \cref{fig:dhist}). 
Therefore, we further add a density regularization $\mathcal{L}_{reg}$ to encourage the densities to approach $1$ so that we can directly set the global threshold to $0.5$:
\begin{equation}
	\label{eq:lreg}
	\mathcal{L}_{reg}=
	\frac{1}{P}
	\sum_{i=1}^P
	\Big(
		\sum_{j=1}^M
		\mathcal{K}(\mathbf{x}_i|\mathbf{g}_j)-1
	\Big)^2
	.
\end{equation}
This loss accelerates training and reduces overlap between primitives by suppressing densities that exceed one, as it facilitates the volume term in $\mathcal{L}_{flm}$.

\subsection{FLM-Occ}
\label{ch3-network}
\begin{figure*}[t]
	\centering
	\includegraphics[width=\textwidth]{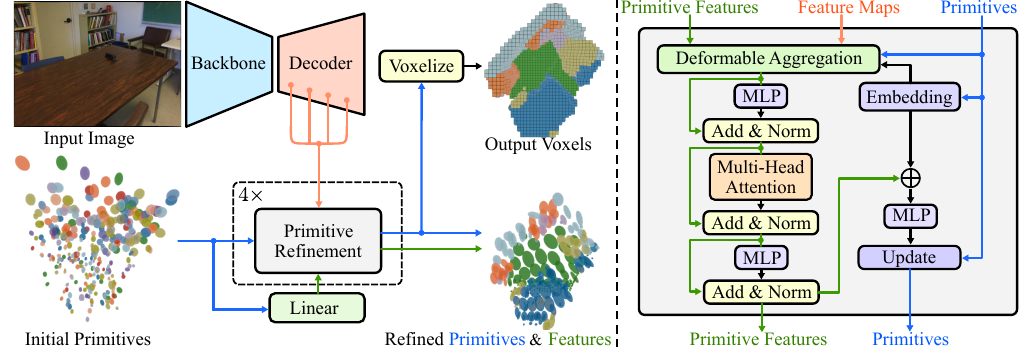}
	\caption{
		\textbf{
			Overview of our FLM-Occ (left) and the structure of the refinement block (right).
		}
		FLM-Occ refines initialized primitives to represent a scene with the guidance of image features.
	}
	\label{fig:pipeline}
\end{figure*}

\textbf{Overview}.
Given an RGB image, camera intrinsics, and a fixed number of uniformly distributed primitives, FLM-Occ gradually relocates and deforms the primitives to match scene structures with the guidance of image features and outputs the corresponding occupancy grid.
As shown in \cref{fig:pipeline}, it consists of an image encoder, a linear layer, a set of learnable primitives, four refinement blocks and a voxelization module.
FLM-Occ features a significantly more streamlined and scalable architecture compared to existing methods~\cite{iso2024eccv,embodiedocc2025iccv,embodiedoccpp2025,splatssc2025}.

\paragraph{Image encoder}
Unlike existing methods, which use Depth Anything v2 (DAv2)~\cite{dav22024nips} and EfficientNet~\cite{effnetv22021pmlr} to obtain geometric and semantic priors, respectively, FLM-Occ relies solely on DAv2 to extract both. 
Our experimental results show that this single-encoder design is sufficient.

\paragraph{Initial primitives}
Each primitive is parameterized by a position $\bm\mu\in\mathbb{R}^3$, scales $\mathbf{s}\in\mathbb{R}^3$, a rotation $\mathbf{R}\in\mathbb{R}^{3\times 3}$, shape parameters $\bm{\epsilon}\in\mathbb{R}^2$ and semantic logits $\mathbf{f}\in\mathbb{R}^n$.
These parameters are learnable but randomly initialized before training.
Notably, this formulation subsumes standard Gaussians as a special case when $\bm{\epsilon}=(1,1)^\top$.
$[\bm{\mu}]_{xy}$ is represented in $[0,1]^2$,
$[\bm{\mu}]_{z}$ and $\mathbf{s}$ are represented in log space.
They are transformed into Euclidean space with camera intrinsics during voxelization.
Primitive features are initialized by applying a linear projection to the primitive parameters.

\paragraph{Refinement block}
Existing methods typically constrain primitive updates to a predefined small range, necessitating careful tuning to balance training stability with the computational overhead of dense voxel grid supervision.
In contrast, leveraging our proposed FLM loss, FLM-Occ eliminates these constraints, enabling unconstrained updates of raw primitive parameters $\tilde{\mbf{g}}$ and $\tilde{\mbf{f}}$:
\begin{equation}
\label{eq:update}
\begin{aligned}
\tilde{\mbf{g}}^{(l+1)}&=\mbf{d}^{(l+1)}_g+\mbf{g}^{(l)}_g,\\
\tilde{\mbf{f}}^{(l+1)}&=\mbf{d}^{(l+1)}_f,
\end{aligned}
\end{equation}
where $[\mbf{d}^{(l+1)}_g;\mbf{d}^{(l+1)}_f]$ is predicted by a multi-layer perceptron (MLP) in the refinement block.
The final primitive parameters are obtained by applying simple activation functions to $\tilde{\mbf{g}}^{(l+1)}$ and $\tilde{\mbf{f}}^{(l+1)}$, including sigmoid for bounded parameters and normalization for quaternions.
Full formulations are provided in the supplementary material.
We use deformable aggregation~\cite{sparse4d2022}, which is commonly used for Gaussian-based occupancy prediction, to sample features for refinement.
We use rotary positional encoding~\cite{su2024roformer} in the multi-head self-attention~\cite{transformer} layers.

\paragraph{Voxelization}
Voxels with densities larger than $0.5$ will be semantically classified with $\operatorname{softmax}$.
We also use neighboring voxelization (\cref{eq:N}) for acceleration.
Notably, calculating training loss in FLM does not require voxelization, and only occupied voxels are used in loss calculation.

\paragraph{Training objective}
FLM-Occ is trained by minimizing the sum of $\mathcal{L}_{flm}$ \big(\cref{eq:flmloss}\big), $\mathcal{L}_{reg}$ \big(\cref{eq:lreg}\big) and a cross entropy loss $\mathcal{L}_{ce}$ for semantics:
\begin{equation}
	\label{eq:loss}
	\min\,\mathcal{L}_{flm}+\lambda_1\mathcal{L}_{reg}+\lambda_2\mathcal{L}_{ce}.
\end{equation}

%% file: tables/ch3-comparison.tex
\begin{table}[tb]
	\centering

	\
	\caption{
		\textbf{Summary of mixture model-based methods.}
		For notational simplicity, the density of voxel $\mathbf{x}$ from the $j$-th  primitive $\mathcal{K}(\mathbf{x}|\mathbf{g}_j)$ is denoted by $\mathcal{K}_j$.
		Normalization for semantic aggregation is omitted for brevity.
		Type 1 treats “empty” as one of the semantic classes, 
		while Type 2 and ours model geometry and semantics separately.
	}
	\resizebox{100mm}{!}{
	\renewcommand{\arraystretch}{1.2}
	\begin{tabular}{l|c|c|c|c}
		\toprule
		\multirow{2}{*}{Method} & \multirow{2}{*}{Type} &

		\multicolumn{2}{c|}{Voxelization Formula} &
		Training\\
		\cline{3-4}
		& &  Geometry & Semantics & Objective  \\
		\midrule

		GaussianFormer~\cite{gaussianformer2024eccv} & 1 & \multicolumn{2}{c|}{$\sum_{j\in\mathcal{N}}a_j\mathcal{K}_j\mathbf{f}_j$} & \multirow{4}{*}{\rotatebox{90}{Voxel Classifcation}} \\[6pt]
		\cline{1-4}
		GaussianFormer-2~\cite{gaussianformer22025cvpr}& &  $1-\prod_{j\in\mathcal{N}}(1-\mathcal{K}_j)$ & 
		$\sum_{j\in\mathcal{N}}\frac{a_j}{v_j}\mathcal{K}_j\mathbf{f}'_j$ & \\[6pt]
		
		QuadricFormer~\cite{quadricformer2025}& 2 & $1-\prod_{j\in\mathcal{N}}(1-\mathcal{K}_j)$ &  
		$\sum_{j\in\mathcal{N}}a_j\mathcal{K}_j\mathbf{f}'_j$ & \\[6pt]
		
		SplatSSC~\cite{splatssc2025}& & $1-\prod_{j\in\mathcal{N}}(1-a_j \mathcal{K}_j)$ &
		$\sum_{j\in\mathcal{N}}\frac{1}{v_j}\mathcal{K}_j\mathbf{f}'_j$ & \\[6pt]
		\midrule
		FLM-Occ (ours) & New& $\sum^M_{j=1}\mathcal{K}_j$     &  $\sum^M_{j=1}\mathcal{K}_j\mathbf{f}_j$  & MLE\\
		\bottomrule

	\end{tabular}
	\label{tb:mm}
	}

\end{table}

%% file: sec/4_exp.tex
\section{Experiments}
\subsection{Dataset and Metrics}
Following prior work, we evaluate our method on indoor datasets Occ-ScanNet~\cite{iso2024eccv} and Occ-ScanNet-mini2~\cite{embodiedocc2025iccv}.
Their train and test splits are $45610/19705$ and $5504/2376$, respectively.
Each sample includes an RGB image, camera intrinsics, and a $60\times60\times36$ ground-truth voxel grid with $11$ semantic classes and a voxel size of $\SI{0.08}{m}$.  
We report the intersection-over-union (IoU) for geometry (occupied vs. empty) and for each semantic class, as well as the mean IoU (mIoU) across all classes, all in percentages (\%).

\subsection{Training Details}
The loss weights $\lambda_1$ and $\lambda_2$ are set to $0.1$ and $1.0$, respectively.  
The model is optimized using AdamW~\cite{adamw2019iclr} with $\beta_1=0.85$ and $\beta_2=0.95$. 
The weight decay is set to $0.01$.
The learning rate is initialized to $\num{5e-4}$ with a 1,000-iteration linear warmup and cosine annealing schedule, whereas the ViT of DAv2~\cite{dav22024nips} is fully fine-tuned with a learning rate of $\num{5e-5}$.
We train FLM-Occ with \texttt{BFloat16} for 30{,}000 iterations on Occ-ScanNet~\cite{iso2024eccv} and 15{,}000 iterations on Occ-ScanNet-mini2~\cite{embodiedocc2025iccv}, respectively, on four RTX A6000 GPUs with a batch size of 128.
Random horizontal flipping and color jittering are used for data augmentation.

\input{tables/ch4-comparison2}
\subsection{Quantitative Comparison}
We evaluate FLM-Occ and compare it with recent methods~\cite{monoscene2022cvpr, iso2024eccv, embodiedocc2025iccv, embodiedoccpp2025, splatssc2025}. 
As reported in \cref{tab:quan}, FLM-Occ achieves state-of-the-art performance, significantly outperforming existing voxel-based and Gaussian-based methods.
Specifically, with only 1024 superquadrics, FLM-Occ reaches an mIoU of 64.0\% on Occ-ScanNet, representing a substantial improvement of +12.2\% over the previous best-performing method, SplatSSC (51.8\%), despite using fewer primitives.
Morever, when the primitive count is reduced to a mere 32 and 64, FLM-Occ still achieves 55.9\% and 56.3\% mIoU, already surpassing SplatSSC (1200 Gaussians)~\cite{splatssc2025} and EmbodiedOcc (16200 Gaussians)~\cite{embodiedocc2025iccv}.
These results suggest that MLE is naturally well aligned with mixture-model-based occupancy prediction.

In terms of implementation, FLM-Occ offers greater simplicity and flexibility compared to existing methods.
While SplatSSC and EmbodiedOcc rely on a two-stage pipeline to fine-tune a depth estimation branch for Gaussian initialization, FLM-Occ learns the initial primitives in a single end-to-end process.
Furthermore, unlike prior works that utilize EfficientNet~\cite{effnetv22021pmlr} and DAv2~\cite{dav22024nips} to obtain semantic and geometric priors, respectively for guiding the refinement, we employ features from a single DAv2~\cite{dav22024nips} and obtain more accurate results.
This observation is consistent with the findings in DAv2, suggesting that representations learned for depth estimation can generalize effectively to semantic occupancy tasks with fine-tuning.

\subsection{Ablation on Primitives}

\begin{figure*}[t]
	\centering
	\includegraphics[width=\textwidth]{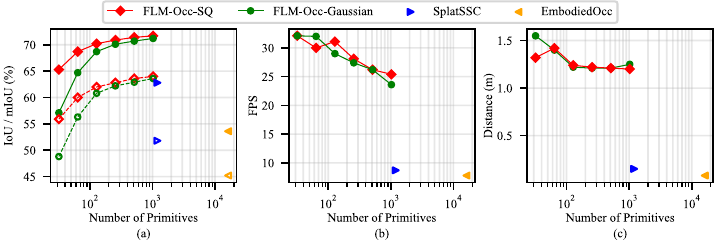}
	\caption{
		\textbf{
			Performance vs. Representation Size.
		}
		Filled and hollow markers in (a) indicate IoU and mIoU, respectively.
		FPS is measured on a single RTX 3090 GPU.
	}
	\label{fig:albation_kernel}
\end{figure*}

We train FLM-Occ with different primitives (Gaussian vs. superquadric) and different numbers of primitives (32,64,128,256,512,1024) to evaluate how primitives affect the results on Occ-ScanNet~\cite{iso2024eccv}.
We also calculate the average distance between the initialized and final primitives.
The results are shown in \cref{fig:albation_kernel}.

Despite using far fewer primitives, FLM-Occ achieves higher accuracy and efficiency than prior SoTA. SplatSSC~\cite{splatssc2025} uses 1,200 Gaussians and achieves 62.8\% IoU and 51.8\% mIoU at 8.7 FPS, whereas our model with only 512 superquadrics reaches 71.4\% IoU and 63.6\% mIoU at 26.2 FPS.
EmbodiedOcc~\cite{embodiedocc2025iccv} relies on 16,200 Gaussians but achieves only 53.6\% IoU and 45.2\% mIoU at 7.8 FPS.
These results demonstrate the efficiency of our FLM formulation.

Increasing the number of primitives improves accuracy while reducing inference speed.
However, the improvement gradually saturates after 256 primitives, suggesting that a few hundred primitives are sufficient to represent the scene.
Superquadrics consistently outperform Gaussians across all primitive counts, especially when less primitives are used.
As the primitive count increases, the performance gap between the Gaussian and superquadric gradually narrows.

The distances between initial and final positions of the primitives generally decrease as the primitive count increases, indicating smaller adjustments during optimization.
The distances (around 1.2 m) of FLM-Occ are significantly larger than those of SplatSSC~\cite{splatssc2025} (0.15 m) and EmbodiedOcc~\cite{embodiedocc2025iccv} (0.08 m), since those methods initialize Gaussians using depth maps or use a large number of primitives, while our primitives are uniformly sampled in the space.
This demonstrates that FLM-Occ learns to substantially relocate the primitives.

\subsection{Ablation Study on Network Configuration}
\label{ch4:ablation}

We utilize FLM-Occ with 128 superquadrics trained on Occ-ScanNet~\cite{iso2024eccv}, as this configuration represents the elbow point of the performance-primitive curve in \cref{fig:albation_kernel}, suggesting it is more sensitive to parameter variations.

In \cref{tab:alba}, the results show that metric depth features improve the performance.
Both IoU and mIoU increases 5 points by replacing the relative DAv2~\cite{dav22024nips} with the metric one (row 1 vs. 2).
This explains why previous methods~\cite{embodiedocc2025iccv,embodiedoccpp2025,splatssc2025} choose to further finetune the metric DAv2~\cite{dinov2} on the ScanNet dataset~\cite{scannet2017cvpr}.
With the ViT~\cite{dinov2} finetuned, the performance significantly improves, especially when using relative DAv2 features (row 1 vs.  3), where the mIoU increases from 43.9\% to 59.2\% (+15.3\%).
This compares to a gain of 10.7\% for metric features (row 2 vs. 4).
These results suggest that while metric-scale features provide a stronger zero-shot prior for 3D occupancy, task-specific finetuning effectively mitigates the scale ambiguity inherent in relative features, allowing them to achieve comparable performance (row 3 vs. 4).
The performance improves with more refinement blocks but saturates at four blocks (rows 5--8).
Furthermore, scaling the encoder to DAv2-ViT-L yields additional gains, achieving higher results of 72.7\% IoU and 65.0\% mIoU (row 8 vs. 9).
The density regularization loss $\mathcal{L}_{reg}$ \big(\cref{eq:lreg}\big) improves the performance (row 4 vs. 7) as we assume that densities lie within $[0,1]$ and set a threshold of $0.5$ for occupancy.

\input{tables/ch4-abla_network}

\begin{figure}[t]
	\centering
	\includegraphics[width=97.6mm]{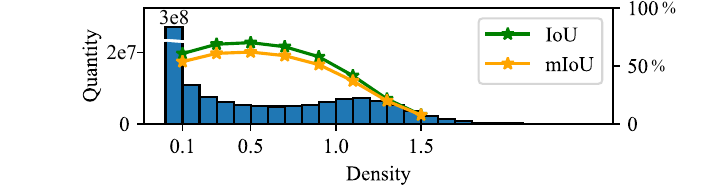}
	\caption{
		\textbf{Distribution of density and IoUs with different density thresholds.}
	}
	\label{fig:dhist}
\end{figure}

\subsection{Influence of Density Threshold for Occupancy}
Although the density regularization loss encourages voxel densities to approach $1$ in occupied regions, some densities still exceed $1$.
Therefore, we analyze the empirical distribution of predicted densities by FLM-Occ with $128$ superquadrics trained on Occ-ScanNet~\cite{scannet2017cvpr} and evaluate how different occupancy thresholds influence IoU and mIoU.
As shown in \cref{fig:dhist}, setting the threshold to $0.5$ yields better results.

\subsection{Qualitative Results}
\begin{figure*}[t]
	\centering
	\includegraphics[width=\textwidth]{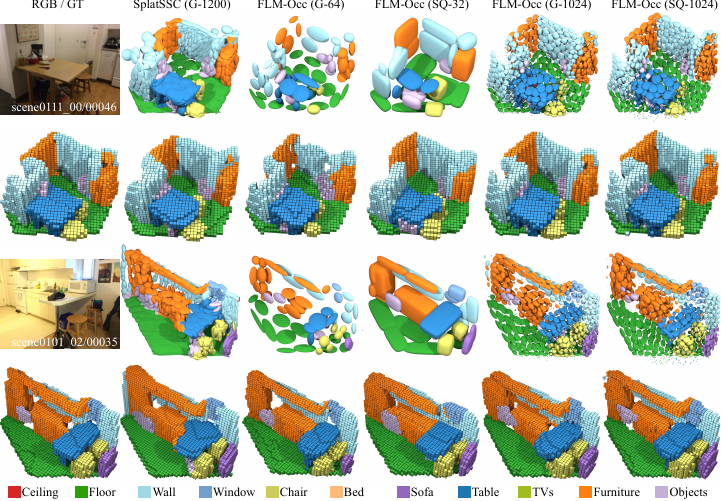}
	\caption{
		\textbf{
			Comparisons on Occ-ScanNet dataset~\cite{iso2024eccv}
		}
	}
	\label{fig:qual}
\end{figure*}

\cref{fig:qual} illustrates the qualitative results of SplatSSC (1200 Gaussians)~\cite{splatssc2025} alongside FLM-Occ under four primitive configurations (64/1024 Gaussians and 32/1024 superquadrics).
Due to the relatively coarse annotation granularity ($0.08$m), all methods effectively capture the scene structure.
However, regarding primitive distribution, SplatSSC exhibits significant inter-primitive overlaps, as its geometric aggregation formula incentivizes such redundancy.
In contrast, FLM-Occ produces sparser representations, a direct result of our MLE-based formulation and density regularization, which encourages a more efficient spatial distribution.

%% file: tables/ch4-comparison2.tex
\definecolor{ceiling}{RGB}{214,  38, 40}   %
\definecolor{floor}{RGB}{43, 160, 4}     %
\definecolor{wall}{RGB}{158, 216, 229}  %
\definecolor{window}{RGB}{114, 158, 206}  %
\definecolor{chair}{RGB}{204, 204, 91}   %
\definecolor{bed}{RGB}{255, 186, 119}  %
\definecolor{sofa}{RGB}{147, 102, 188}  %
\definecolor{table}{RGB}{30, 119, 181}   %
\definecolor{tvs}{RGB}{160, 188, 33}   %
\definecolor{furniture}{RGB}{255, 127, 12}  %
\definecolor{objects}{RGB}{196, 175, 214} %

\begin{table*}[t]
\small
\centering
\setlength{\tabcolsep}{0.85mm}
\caption{
	\textbf{Results on Occ-ScanNet~\cite{iso2024eccv}.}
	Best, second, and third results are highlighted in light red, orange, and yellow, respectively.
	The ``Type'' follows the definitions in \cref{tb:mm}.
	``Vox'', ``Que'', ``Gau'', and ``SQ'' denote voxel, query, Gaussian, and superquadric, respectively.
}
\label{tab:quan}
\resizebox{\textwidth}{!}{
\begin{tabular}{c|l|ccc|c|ccccccccccc|c}
\toprule 
\rotatebox{90}{Dataset}
&Method & Num. & Rep. & Type
& IoU
& \rotatebox{90}{\textcolor{ceiling}{$\blacksquare$} Ceiling} 
& \rotatebox{90}{\textcolor{floor}{$\blacksquare$} Floor}
& \rotatebox{90}{\textcolor{wall}{$\blacksquare$} Wall} 
& \rotatebox{90}{\textcolor{window}{$\blacksquare$} Window} 
& \rotatebox{90}{\textcolor{chair}{$\blacksquare$} Chair} 
& \rotatebox{90}{\textcolor{bed}{$\blacksquare$} Bed} 
& \rotatebox{90}{\textcolor{sofa}{$\blacksquare$} Sofa} 
& \rotatebox{90}{\textcolor{table}{$\blacksquare$} Table} 
& \rotatebox{90}{\textcolor{tvs}{$\blacksquare$} TVs} 
& \rotatebox{90}{\textcolor{furniture}{$\blacksquare$} Furniture} 
& \rotatebox{90}{\textcolor{objects}{$\blacksquare$} Objects} 
& mIoU\\

\midrule
\multirow{11}{*}{\rotatebox{90}{Occ-ScanNet}}
&MonoScene~\cite{monoscene2022cvpr} & 129.6k & Vox & --- & 41.6 & 15.2 & 44.7 & 22.4 & 12.6 & 26.1 & 27.0 & 35.9 & 28.3 & \ 6.6 & 32.2 & 19.8 & 24.6 \\

&ISO~\cite{iso2024eccv} & 129.6k & Vox & --- & 42.2 & 19.9 & 41.9 & 22.4 & 17.0 & 29.0 & 42.4 & 42.0 & 29.6 & 10.6 & 36.4 & 24.6 & 28.7 \\

&DiScene~\cite{discene2025ral} & 960 & Que & --- & 47.2 & 45.3 & 50.6 & 40.4 & 36.7 & 42.3 & 59.7 & 62.0 &45.6 & 41.2 & 52.4 & 42.7 & 47.2 \\

&E.Occ~\cite{embodiedocc2025iccv} & 16200 & Gau & 1 & 53.6 & 39.6 & 50.4 & 41.4 & 31.7 & 40.9 & 55.0 & 61.4 & 44.0 & 36.1 & 53.9 & 42.2 & 45.2 \\ 

&E.Occ++~\cite{embodiedoccpp2025} & 16200 & Gau & 1 & 54.9 & 36.4 & 53.1 & 41.8 & 34.4 & 42.9 & 57.3 & 64.1 & 45.2 & 34.8 & 54.2 & 44.1 & 46.2 \\


&SplatSSC~\cite{splatssc2025} & 1200 & Gau & 2 & 62.8 & 49.1 & 59.0 & 48.3 & 38.8 & 47.4 & 62.4 & 67.0
& 49.5 & 42.6 & 60.7 & 45.4 & 51.8 \\

\cmidrule(lr){2-18}

&\multirow{4}{*}{FLM-Occ}

& 64 & Gau & \multirow{4}{*}{New}
& 64.7 & 55.0 & 64.0 & 52.0 & \third{44.7} & \third{48.6}
& \third{67.3} & 70.4 & \third{54.3} & \third{46.7} & 64.3
& \third{52.4} & \third{56.3} \\

&& 32 & SQ &
& \third{65.3} & \third{55.1} & \third{64.5} & \third{52.2} & 43.3 & 48.3
& 66.5 & \third{70.6} & 53.5 & 45.6 & \third{64.4}
& 50.7 & 55.9 \\

&& 1024 & Gau &
& \second{71.2} & \second{61.9} & \second{70.0} & \second{57.5} & \second{52.6} & \second{58.1}
& \second{73.5} & \second{77.5} & \second{62.7} & \best{55.5} & \second{70.3}
& \second{59.8} & \second{63.6} \\

&& 1024 & SQ &
& \best{71.7} & \best{62.5} & \best{70.6} & \best{58.0} & \best{53.1} & \best{58.8}
& \best{74.0} & \best{77.9} & \best{63.5} & \second{55.2} & \best{70.8}
& \best{60.3} & \best{64.0} \\

\midrule

\multirow{10}{*}{\rotatebox{90}{Occ-ScanNet-mini2}}

&MonoScene~\cite{monoscene2022cvpr} & 129.6k & Vox & --- 
& 41.9  
& 17.0 & 46.2 & 23.9 & 12.7 & 27.0 & 29.1 & 34.8 & 29.1 & \ 9.7 & 34.5 & 20.4 
& 25.9 \\

&ISO~\cite{iso2024eccv} & 129.6k & Vox & --- 
& 42.9  
& 21.1 & 42.7 & 24.6 & 15.1 & 30.8 & 41.0 & 43.3 & 32.2 & 12.1 & 35.9 & 25.1 
& 29.4 \\

&E.Occ~\cite{embodiedocc2025iccv} & 16200 & Gau & 1 
& 55.1  
& 29.5 & 49.4 & 41.7 & 36.3 & 41.9 & 60.4 & 59.6 & 46.3 & 34.5 & 58.0 & 43.5 
& 45.6 \\

&E.Occ++~\cite{embodiedoccpp2025} & 16200 & Gau & 1 
& 55.7  
& 23.3 & 51.0 & 42.8 & 39.3 & 43.5 & 65.6 & 64.0 & 50.7 & \third{40.7} & 60.3 & 48.9 
& 48.2 \\

&SplatSSC~\cite{splatssc2025} & 1200 & Gau & 2 
& 61.5  
& 36.6 & 55.7 & 46.5 & 40.1 & 45.6 & 64.5 & 62.4 & 48.6 & 30.6 & 61.2 & 45.4 
& 48.9 \\

\cmidrule(lr){2-18}

&\multirow{4}{*}{FLM-Occ}

& 64 & Gau  & \multirow{4}{*}{New} &
62.1 & 42.4 & 62.1 & 49.3 & 45.5 & 46.8 & 67.4 & 65.9 & 55.3 & 38.4 & 66.3 & 50.9 & 53.6 \\

&& 32 & SQ & &
\third{65.3} & \third{45.8} & \third{64.8} & \third{51.7} & \third{47.8} & \third{49.0} & \third{67.8} & \third{67.8} & \third{57.6} & 38.9 & \third{66.7} & \third{52.0} & \third{55.4} \\

&& 1024 & Gau & & 
\second{72.6} & \second{56.9} & \second{72.7} & \second{56.9} & \second{57.5} & \second{62.3} & \second{76.5} & \second{77.7} & \best{70.4} & \second{49.3} & \second{75.0} & \second{62.6}  & \second{65.3} \\

&& 1024 & SQ & & 
\best{72.9} & \best{57.9} & \best{72.8} & \best{57.4} & \best{58.0} & \best{63.1} & \best{76.7} & \best{77.8} & \best{70.4} & \best{51.0} & \best{75.3} & \best{62.8} & \best{65.7} \\

\bottomrule
\end{tabular}
}

\end{table*}

%% file: tables/ch4-abla_network.tex
\begin{table}[t]
	\centering
	\small
	\setlength{\tabcolsep}{1.5mm}
	\caption{
		\textbf{Performance with different network configurations.}
	}
	\label{tab:alba}
	\resizebox{100mm}{!}{
		\begin{tabular}{c|c|c|clc|c|cc}
			\toprule
			&SQ Count &Refine& DAv2~\cite{dav22024nips} & ViT~\cite{dinov2} & Fintune ViT & $\mathcal{L}_{reg}$ & IoU & mIoU \\
			\midrule
			
			1)&\multirow{9}{*}{128}
			&4&Rel &  Base &     &           & 52.4  & 43.9     \\
			
			2)&&4& Metric &  Base  &    &     & 57.3 & 48.9 \\
			
			3)&&4& Rel &  Base  & \checkmark    &           & 67.2  & 59.2   \\
			
			4)&&4& Metric & Base & \checkmark    &     & 67.5 & 59.6 \\
			\cline{3-9}
			5)&&2& Metric & Base & \checkmark    & \checkmark & 69.2 & 60.7 \\
			
			6)&&3& Metric & Base & \checkmark    & \checkmark & 69.8 & 61.5 \\
			
			7)&&4& Metric & Base & \checkmark    & \checkmark & 70.2 & 62.0 \\
			
			8)&&5& Metric & Base & \checkmark    & \checkmark & 70.0 & 61.8 \\
			\cline{3-9}
			9)&&\tbf{4}&  \tbf{Rel} &  \tbf{Large} &  \checkmark    & \checkmark & \tbf{72.7} & \tbf{65.0} \\
			\midrule
			10)&1024 &4& Metric & Base & \checkmark    & \checkmark & 71.7 & 64.0 \\
			\bottomrule
		\end{tabular}
	}
\end{table}

%% file: sec/5_conclusion.tex
\section{Conclusion}
We propose FLM-Occ---Feed-forward Likelihood Maximization for efficient indoor occupancy prediction---a novel method that reformulates occupancy prediction not as voxel classification but as voxel distribution estimation, parameterized by a mixture model.
To leverage MLE for end-to-end training, we propose FLM loss, a novel loss that explicitly drives the predicted primitives to match objects.
We also propose probabilistic voxelization formulas to convert a mixture model to an occupancy grid.
Built upon FLM, FLM-Occ is able to substantially relocate and deform geometric primitives to represent a scene.
Compared to the prior state-of-the-art method, FLM-Occ leverages 20$\times$ fewer primitives to achieve higher accuracy and over 3$\times$ speed up on Occ-ScanNet datasets.

%% file: sec/acknowl.tex
\\
\\
\noindent\textbf{Acknowledgements.}
This work was supported in part by Shenzhen Science and Technology Program (No. SGDX20240115111759002), in part by Meituan Academy of Robotics Shenzhen, in part by the Shenzhen Association for Science and Technology (No. XHXS2025-003), in part by high level of special funds (G03034K003) from Southern University of Science and Technology, Shenzhen, China, and in part by the Open Research Fund (GML-KF-24-15) from Guangdong Laboratory of Artificial Intelligence and Digital Economy.

%% file: sec/X_suppl.tex
\begin{center}
	{\Large\bfseries Supplementary Material}
\end{center}

\section{Discussion}
\subsection{Limitations}
\paragraph{Bilinear Complexity of the FLM Loss}
The computational cost of the FLM loss scales with the number of occupied voxels and primitives, which limits its applicability to large-scale scenes.
Incorporate local and global aggregations within a hierarchical structure of primitives should reduce the cost.

\paragraph{Fixed Number of Primitives}
Similar to EmbodiedOcc~\cite{embodiedocc2025iccv} and SplatSSC ~\cite{splatssc2025}, FLM-Occ relies on a fixed number of primitives, meaning that changing the primitive count requires retraining the network.
One potential direction to address this limitation is to incorporate learnable pruning and splitting mechanisms, similar to those adopted in sparse voxel-based methods~\cite{octmae2024eccv,octreeocc2024nips}.

\paragraph{Limited Dataset Scope}
Following protocols used by EmbodiedOcc~\cite{embodiedocc2025iccv} and SplatSSC~\cite{splatssc2025}, we evaluate our method on Occ-ScanNet~\cite{iso2024eccv} and its subset, Occ-ScanNet-mini2~\cite{embodiedocc2025iccv}.
However, the annotation quality and granularity of Occ-ScanNet are relatively limited, as its labeling pipeline inherits from the legacy NYUv2~\cite{nyueccv2012} dataset.
We believe that higher-quality datasets with finer annotations would help further explore and unleash the potential of FLM-Occ.
Besides, our FLM loss should be applicable to outdoor settings, since its formulation is inherently independent of scene type.

We will address these limitations in future work.

\subsection{Future Work and Broader Applications}
\paragraph{Future work}
Although this work focuses on indoor occupancy prediction, we plan to extend FLM-Occ to autonomous driving scenarios, leveraging its scene-agnostic architecture.
Unlike prior methods, FLM-Occ does not impose constraints on the step size of primitive refinements, which typically requires meticulous parameter tuning across varying scene scales and voxel resolutions~\cite{embodiedocc2025iccv,splatssc2025}.
This flexibility significantly enhances the feasibility of such extensions.
We are also interested in training FLM-Occ with RGB-D sequences that are more readily available than ground-truth voxel grids, especially given the current scarcity of high-quality indoor occupancy datasets.

\paragraph{Broader applications}
The FLM loss facilitates direct supervision of primitives via point clouds, and FLM-Occ showcases how this loss can optimize parameters to achieve a more efficient parameterization of 3D scenes.
Beyond occupancy prediction,
we believe the FLM loss can also be applied to tasks that predicts 3D primitives, such as feed-forward Gaussian splatting~\cite{gslrm2024eccv,noposplat2025iccv}.
Another interesting direction is to leverage the FLM loss to supervise explicit primitives that define the spatial extent of a neural field, thereby achieving a hybrid representation that is both expressive and efficient for complex 3D scenes.

\section{Toy Experiments on the Loss Functions}
\begin{figure}[t]
	\centering
	\includegraphics[width=122mm]{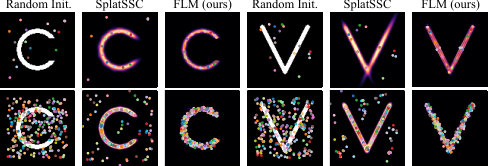}
	\caption{
		\textbf{
			Fitting shapes with 16 / 256 Gaussians using two losses, respectively.
		}
		The 1st and 4th columns show the target shapes and initializations.
		Only the positions of the Gaussians are visualized.
		Bright pixels indicate occupied regions.
	}
	\label{fig:two_losses}
\end{figure}

To show that our FLM loss is robust to random initialization rather than relying on good initializations required by SplatSSC~\cite{splatssc2025},
we conduct a 2D toy experiment to isolate the effect of different training objectives.
The results are shown in \cref{fig:two_losses}.
Although the loss used by SplatSSC can also move some Gaussians over long distances, it leaves some Gaussians isolated in empty regions, as indicated by the gradient analysis in \cref{eq:gf2grad}.
This explains why SplatSSC relies on depth maps for initializing the Gaussians.
In contrast, FLM loss encourages all Gaussians to move toward and explain occupied points, enabling more effective relocation from random initialization.

\begin{figure*}[!t]
	\centering
	\includegraphics[width=122mm]{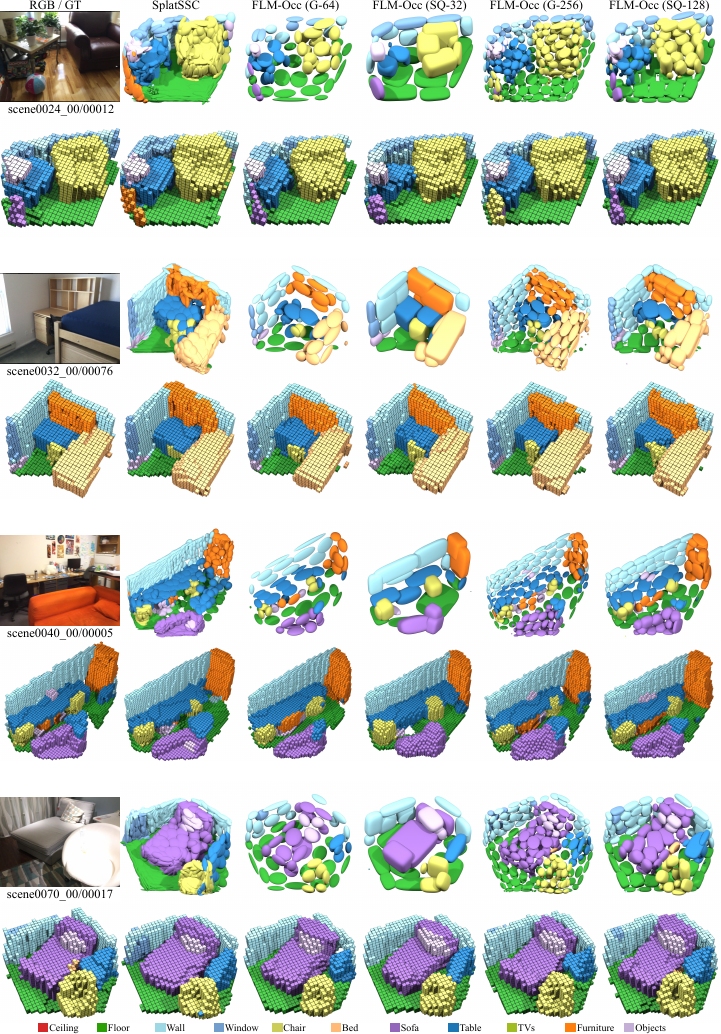}
	\caption{
		\textbf{
			Predicted primitives and voxels by SplatSSC~\cite{splatssc2025} and FLM-Occ.
		}
	}
	\label{fig:qual2}
\end{figure*}

\section{Primitive Parameter Activation}
\label{supp:imple}

Given the raw primitive parameters $\tilde{\mbf{g}}^{(l+1)}$ and $\tilde{\mbf{f}}^{(l+1)}$ predicted by the update head of the refinement block, the primitive parameters are calculated by:
\begin{equation}
	\label{eq:activate}
	\begin{aligned}
		[\bm{\mu}^{(l+1)}]_{xy}
		&= \mathrm{Sigmoid}\big([\tilde{\mbf{g}}^{(l+1)}]_{xy}\big), \\
		[\bm{\mu}^{(l+1)}]_{z}
		&= [\tilde{\mbf{g}}^{(l+1)}]_{z}, \\
		\mathbf{s}^{(l+1)}
		&= [\tilde{\mbf{g}}^{(l+1)}]_{s}, \\
		\mathbf{q}^{(l+1)}
		&= \mathrm{Normalize}([\tilde{\mbf{g}}^{(l+1)}]_{q}), \\				\bm{\epsilon}^{(l+1)}
		&= \mathrm{Sigmoid}\big([\tilde{\mbf{g}}^{(l+1)}]_{\epsilon}\big), \\
		\mathbf{f}^{(l+1)}
		&=\tilde{\mbf{f}}^{(l+1)}.
	\end{aligned}
\end{equation}
The scale in Euclidean space is computed as $\exp(\mathbf{s} + s_{\min})$, where $s_{\min}$ is introduced to prevent the kernels from collapsing to excessively small scales in empty regions.
In our experiments, $s_{\min}$ is set to $0.02$ when the primitive count is below $1024$, and reduced to $0.01$ when the count reaches $1024$.
A smaller $s_{\min}$ is adopted because a larger number of primitives with excessive minimum volumes may cause their aggregate volume to exceed the actual occupancy, leading to occupancy errors.

\section{Raw Data of the Ablation Study on Primitive}
\input{tables/ch4-kernel_count}

The values used in the ablation study on primitive in the manuscript are reported in \cref{tab:kernel}.
``Dists.'' denotes the average distance between initial positions and final positions of the primitives.
Note that the FPS values are measured using \texttt{bfloat16} on a single RTX 3090 GPU.

\section{Additional Qualitative Results}
More qualitative results of FLM-Occ with SplatSSC~\cite{splatssc2025} (using 1,200 Gaussians) are presented in \cref{fig:qual2}.
The configuration for FLM-Occ includes four primitive settings: 64/256 Gaussians and 32/128 superquadrics.
We omit results for higher primitive counts, as performance saturates beyond these configurations.

SplatSSC fails to predict large-scale primitives for simple, expansive structures due to its voxel-classification training objective and constraints on primitive refinement. Furthermore, its multiplicative aggregation formula encourages excessive overlapping among primitives, which hinders representational efficiency.
In contrast, FLM-Occ achieves a sparse, nearly object-level representation and incrementally captures finer details as the primitive count increases.

%% file: tables/ch4-kernel_count.tex
\begin{table}[t]
	\centering
	\small
	\setlength{\tabcolsep}{1.6mm}
	\caption{\textbf{Ablation study on kernel function and primitive count.}}
	\label{tab:kernel}
	\resizebox{100mm}{!}{%
	\begin{tabular}{r|rcrc|rcrc}
		\toprule
		\multirow{2}{*}{Number}
		& \multicolumn{4}{c|}{Gaussian}
		& \multicolumn{4}{c}{Superquadric} \\
		\cmidrule(lr){2-5}\cmidrule(lr){6-9}
		& IoU & mIoU & FPS & Dist. (m) & IoU & mIoU & FPS & Dist. (m) \\
		\midrule
		32   & 57.1 & 48.8 & 32.1 & 1.55 & 65.3 & 55.9 & 32.1 & 1.32\\
		64   & 64.7 & 56.3 & 32.0 & 1.40 & 68.7 & 60.0 & 30.0 & 1.42\\
		128   & 68.7 & 60.8 & 29.0 & 1.22 & 70.2 & 62.0 & 31.1 & 1.24\\
		256   & 70.1 & 62.2 & 27.4 & 1.21 & 70.9 & 62.8 & 28.1 & 1.22\\
		512   & 70.7 & 62.9 & 26.2 & 1.21 & 71.4 & 63.6 & 26.2 & 1.21\\
		1024   & 71.2 & 63.6 & 23.6 & 1.25 & 71.7 & 64.0 & 25.4 & 1.20\\
		\bottomrule
	\end{tabular}
	}
\end{table}

%% file: sec/X_kernel_volume.tex
\section{Derivation of the Closed-Form Expression for Superquadric Volume}
We use superquadric formulas to unify the closed forms of kernel volume $v_j$ for superquadric and Gaussian.
Each superquadric $\mathbf{k}$ has a position $\mathbf{x}=(x,y,z)^\top$, a rotation $\mathbf{R}$, scales $\mathbf{s}=(s_x,s_y,s_z)^\top$ and shape parameters $\bm{\epsilon}=(\epsilon_1,\epsilon_2)^\top$.
We define the superquadric kernel function as follows:
\begin{gather}
	\label{eq:Ksq}
	\mathcal{K}_q(\mathbf{x},\mathbf{k})=
	\exp\big(-\frac{1}{2}f(\mathbf{x},\mathbf{k})\big)
	,
	\\
	\label{eq:sq}
	f(\mathbf{x},\mathbf{k})=
	\Big(\big(\frac{x}{s_x}\big)^{\frac{2}{\epsilon_2}}+
	\big(\frac{y}{s_y}\big)^{\frac{2}{\epsilon_2}}
	\Big)^{\frac{\epsilon_2}{\epsilon_1}}
	+\big(\frac{z}{s_z}\big)^{\frac{2}{\epsilon_1}}.
\end{gather}
Note that $\mathbf{x}$ should be transformed into the frame of the superquadric using $\mathbf{R}^\top \mathbf{x}$.
When $\epsilon_1=\epsilon_2=1.0$, the above expressions reduce to those of the Gaussian.
The volume $v_g$ of a Gaussian is as simple as
\begin{gather}
	\label{eq:v_g}
	v_g=(2\pi)^{\frac{3}{2}}|\bm\Sigma|^{\frac{1}{2}}
	=(2\pi)^{\frac{3}{2}}s_xs_ys_z.
\end{gather}

Now we show the integral for the volume of a superquadric.
Since the integrand is an even function w.r.t. $x$, $y$, and $z$, we use the symmetry of a superquadric to reduce the region to the first octant:
\begin{equation}
	\label{eq:integral}
	I = 8\int_{x,y,z\ge0}
	\exp\big(-\frac{1}{2}f(\mathbf{x},\mathbf{k})\big)
	\,dx\,dy\,dz.
\end{equation}
We introduce the change of variables
\begin{equation}
	u=\Big(\frac{x}{s_x}\Big)^{\tfrac{2}{\epsilon_2}},\ 
	v=\Big(\frac{y}{s_y}\Big)^{\tfrac{2}{\epsilon_2}},\ 
	w=\Big(\frac{z}{s_z}\Big)^{\tfrac{2}{\epsilon_1}}.
\end{equation}
and then obtain
\begin{equation}
\begin{aligned}
	dx &= s_x\frac{\epsilon_2}{2}\,u^{\frac{\epsilon_2}{2}-1}\,du,\\
	dy &= s_y\frac{\epsilon_2}{2}\,v^{\frac{\epsilon_2}{2}-1}\,dv,\\
	dz &= s_z\frac{\epsilon_1}{2}\,w^{\frac{\epsilon_1}{2}-1}\,dw.
\end{aligned}
\end{equation}
Substituting the above equations into \cref{eq:integral}, we get
\begin{multline}
	I = 8\,s_x s_y s_z \left(\tfrac{\epsilon_2}{2}\right)^2\left(\tfrac{\epsilon_1}{2}\right)
	\int_{0}^{\infty}\!\!\int_{0}^{\infty}\!\!\int_{0}^{\infty}\\
	u^{a-1} v^{a-1} w^{b-1} e^{-\tfrac{1}{2}((u+v)^{c}+w)}\,du\,dv\,dw,
\end{multline}
where $a=\tfrac{\epsilon_2}{2},\ b=\tfrac{\epsilon_1}{2},\ c=\tfrac{\epsilon_2}{\epsilon_1}$.
Using the gamma function $\Gamma(z)=\int^{\infty}_0 e^{-t}t^{z-1}dt$~\cite{gamma}, we first integrate w.r.t. $w$:
\begin{equation}
	\label{eq:w}
	\int_0^\infty w^{b-1} e^{-\tfrac{1}{2}w}\,dw = 2^{b}\Gamma(b)=2^{\frac{\epsilon_2}{2}}\Gamma(\tfrac{\epsilon_2}{2}).
\end{equation}
Next, we use the change of variables $t=u+v,\ s=u/(u+v)$ and derive the jacobian
\begin{equation}
	\mathbf{J}(s,t)=
	\begin{bmatrix}
		\frac{\partial u}{\partial s} & \frac{\partial u}{\partial t} \\[1.2ex]
		\frac{\partial v}{\partial s} & \frac{\partial v}{\partial t}
	\end{bmatrix}
	=
	\begin{bmatrix}
		t & s \\
		- t & 1 - s
	\end{bmatrix}.
\end{equation}
Using $dudv=|\mathbf{J}(s,t)|dsdt=tdsdt$, we integrate w.r.t. $u$ and $v$:
\begin{flalign}
	&\int_0^\infty\int_0^\infty u^{a-1} v^{a-1} e^{-\tfrac{1}{2}(u+v)^c}\,du\,dv
	\nonumber\\
	&=\int_0^\infty \int_0^1 (ts)^{a-1}(t(1-s))^{a-1} t\,ds\,dt \; e^{-\tfrac{1}{2}t^c}
	\\
	&= \left(\int_0^1 s^{a-1}(1-s)^{a-1}\,ds\right)
	\left(\int_0^\infty t^{2a-1} e^{-\tfrac{1}{2}t^c}\,dt\right).\nonumber
\end{flalign}
We can compute the integral over $t$ with the change of variables $y=\tfrac{1}{2}t^c$, i.e., $t=(2y)^{1/c}$:
\begin{flalign}
	\label{eq:t}
	&\int_0^\infty t^{2a-1} e^{-\tfrac{1}{2}t^c}\,dt\nonumber\\
	&=\int_0^\infty \big((2y)^{1/c}\big)^{2a-1} e^{-y}\,\frac{2}{c}(2y)^{\frac{1}{c}-1}dy\nonumber\\
	&=\frac{1}{c}2^{\frac{2a}{c}}\int_0^\infty e^{-y}y^{\frac{2a}{c}-1}dy\\
	&=\frac{1}{c}2^{\frac{2a}{c}}\Gamma(\tfrac{2a}{c}).\nonumber
\end{flalign}
Using the Beta function $B(a,b)=\int_0^1t^{a-1}(1-t)^{b-1}dt=\Gamma(a)\Gamma(b)/\Gamma(a+b)$~\cite{gamma},
we compute the integral over $s$:
\begin{equation}
	\label{eq:s}
	\int_0^1 s^{a-1}(1-s)^{a-1}\,ds = B(a,a)
	=\frac{\Gamma(\frac{\epsilon_2}{2})^2}{\Gamma(\epsilon_2)}.
\end{equation}
Combining \cref{eq:w}, \cref{eq:t} and \cref{eq:s}, we get the volume $v_q$ of a superquadric
\begin{flalign}
	\label{eq:v_q}
	v_q &= 8\,s_x s_y s_z \left(\tfrac{\epsilon_2}{2}\right)^2\left(\tfrac{\epsilon_1}{2}\right)
	2^{\frac{\epsilon_2}{2}}\Gamma(\tfrac{\epsilon_2}{2})
	\frac{1}{c}2^{\frac{2a}{c}}\Gamma(\tfrac{2a}{c})
	\frac{\Gamma(\frac{\epsilon_2}{2})^2}{\Gamma(\epsilon_2)}\\
	&=2^{\tfrac{3\epsilon_1}{2}}\;\epsilon_1^2\,\epsilon_2\; 
	\frac{\Gamma\!\big(\tfrac{\epsilon_2}{2}\big)^2}{\Gamma(\epsilon_2)}\;
	\Gamma(\epsilon_1)\;\Gamma\!\big(\tfrac{\epsilon_1}{2}\big)s_x s_y s_z.
\end{flalign}

%% file: main.bbl
\begin{thebibliography}{10}
\providecommand{\url}[1]{\texttt{#1}}
\providecommand{\urlprefix}{URL }
\providecommand{\doi}[1]{https://doi.org/#1}

\bibitem{monoscene2022cvpr}
Cao, A.Q., De~Charette, R.: Monoscene: Monocular 3d semantic scene completion. In: Proceedings of the IEEE/CVF Conference on Computer Vision and Pattern Recognition. pp. 3991--4001 (2022)

\bibitem{pixelsplat2024cvpr}
Charatan, D., Li, S.L., Tagliasacchi, A., Sitzmann, V.: pixelsplat: 3d gaussian splats from image pairs for scalable generalizable 3d reconstruction. In: Proceedings of the IEEE/CVF conference on computer vision and pattern recognition. pp. 19457--19467 (2024)

\bibitem{mvsplat2024eccv}
Chen, Y., Xu, H., Zheng, C., Zhuang, B., Pollefeys, M., Geiger, A., Cham, T.J., Cai, J.: Mvsplat: Efficient 3d gaussian splatting from sparse multi-view images. In: European Conference on Computer Vision. pp. 370--386. Springer (2024)

\bibitem{scannet2017cvpr}
Dai, A., Chang, A.X., Savva, M., Halber, M., Funkhouser, T., Nie{\ss}ner, M.: Scannet: Richly-annotated 3d reconstructions of indoor scenes. In: Proceedings of the IEEE conference on computer vision and pattern recognition. pp. 5828--5839 (2017)

\bibitem{gaussianocc2025iccv}
Gan, W., Liu, F., Xu, H., Mo, N., Yokoya, N.: Gaussianocc: Fully self-supervised and efficient 3d occupancy estimation with gaussian splatting. In: Proceedings of the IEEE/CVF International Conference on Computer Vision. pp. 28980--28990 (2025)

\bibitem{fastocc2024icra}
Hou, J., Li, X., Guan, W., Zhang, G., Feng, D., Du, Y., Xue, X., Pu, J.: Fastocc: Accelerating 3d occupancy prediction by fusing the 2d bird’s-eye view and perspective view. In: 2024 IEEE International Conference on Robotics and Automation (ICRA). pp. 16425--16431. IEEE (2024)

\bibitem{noposplat2025iccv}
Huang, R., Mikolajczyk, K.: No pose at all: Self-supervised pose-free 3d gaussian splatting from sparse views. In: Proceedings of the IEEE/CVF International Conference on Computer Vision. pp. 27947--27957 (2025)

\bibitem{gaussianformer22025cvpr}
Huang, Y., Thammatadatrakoon, A., Zheng, W., Zhang, Y., Du, D., Lu, J.: Gaussianformer-2: Probabilistic gaussian superposition for efficient 3d occupancy prediction. In: Proceedings of the IEEE/CVF Conference on Computer Vision and Pattern Recognition (CVPR). pp. 27477--27486 (June 2025)

\bibitem{selfocc2024cvpr}
Huang, Y., Zheng, W., Zhang, B., Zhou, J., Lu, J.: { SelfOcc: Self-Supervised Vision-Based 3D Occupancy Prediction }. In: 2024 IEEE/CVF Conference on Computer Vision and Pattern Recognition (CVPR). pp. 19946--19956. IEEE Computer Society, Los Alamitos, CA, USA (Jun 2024). \doi{10.1109/CVPR52733.2024.01885}, \url{https://doi.ieeecomputersociety.org/10.1109/CVPR52733.2024.01885}

\bibitem{tpvformer2023cvpr}
Huang, Y., Zheng, W., Zhang, Y., Zhou, J., Lu, J.: Tri-perspective view for vision-based 3d semantic occupancy prediction. In: Proceedings of the IEEE/CVF conference on computer vision and pattern recognition. pp. 9223--9232 (2023)

\bibitem{gaussianformer2024eccv}
Huang, Y., Zheng, W., Zhang, Y., Zhou, J., Lu, J.: Gaussianformer: Scene as gaussians for vision-based 3d semantic occupancy prediction. In: Leonardis, A., Ricci, E., Roth, S., Russakovsky, O., Sattler, T., Varol, G. (eds.) Computer Vision -- ECCV 2024. pp. 376--393. Springer Nature Switzerland, Cham (2025)

\bibitem{octmae2024eccv}
Iwase, S., Liu, K., Guizilini, V., Gaidon, A., Kitani, K., Ambru{\c{s}}, R., Zakharov, S.: Zero-shot multi-object scene completion. In: European Conference on Computer Vision. pp. 96--113. Springer (2024)

\bibitem{scenedino2025iccv}
Jevti{\'c}, A., Reich, C., Wimbauer, F., Hahn, O., Rupprecht, C., Roth, S., Cremers, D.: Feed-forward {SceneDINO} for unsupervised semantic scene completion. In: IEEE/CVF International Conference on Computer Vision (ICCV) (2025)

\bibitem{3dgs2023sig}
Kerbl, B., Kopanas, G., Leimk{\"u}hler, T., Drettakis, G.: 3d gaussian splatting for real-time radiance field rendering. ACM Trans. Graph.  \textbf{42}(4),  139--1 (2023)

\bibitem{occmamba2025cvpr}
Li, H., Hou, Y., Xing, X., Ma, Y., Sun, X., Zhang, Y.: Occmamba: Semantic occupancy prediction with state space models. In: Proceedings of the Computer Vision and Pattern Recognition Conference. pp. 11949--11959 (2025)

\bibitem{sliceocc2025icra}
Li, J., Lu, M., Liu, J., Wang, H., Gu, C., Zheng, W., Du, L., Zhang, S.: Sliceocc: Indoor 3d semantic occupancy prediction with vertical slice representation. In: 2025 IEEE International Conference on Robotics and Automation (ICRA). pp. 15762--15768. IEEE (2025)

\bibitem{discene2025ral}
Li, X., Zheng, Y., Li, P., Chen, Y., Zhang, Y.Q., Ding, W.: Enhancing indoor occupancy prediction via sparse query-based multi-level consistent knowledge distillation. IEEE Robotics and Automation Letters  \textbf{10}(11),  11690--11697 (2025). \doi{10.1109/LRA.2025.3615532}

\bibitem{voxformer2023cvpr}
Li, Y., Yu, Z., Choy, C., Xiao, C., Alvarez, J.M., Fidler, S., Feng, C., Anandkumar, A.: Voxformer: Sparse voxel transformer for camera-based 3d semantic scene completion. In: Proceedings of the IEEE/CVF Conference on Computer Vision and Pattern Recognition (CVPR) (2023)

\bibitem{sparse4d2022}
Lin, X., Lin, T., Pei, Z., Huang, L., Su, Z.: Sparse4d: Multi-view 3d object detection with sparse spatial-temporal fusion (2022)

\bibitem{adamw2019iclr}
Loshchilov, I., Hutter, F.: Decoupled weight decay regularization. In: International Conference on Learning Representations (2019)

\bibitem{octreeocc2024nips}
Lu, Y., Zhu, X., Wang, T., Ma, Y.: Octreeocc: Efficient and multi-granularity occupancy prediction using octree queries. Advances in Neural Information Processing Systems  \textbf{37},  79618--79641 (2024)

\bibitem{monogs2024cvpr}
Matsuki, H., Murai, R., Kelly, P.H., Davison, A.J.: Gaussian splatting slam. In: Proceedings of the IEEE/CVF Conference on Computer Vision and Pattern Recognition. pp. 18039--18048 (2024)

\bibitem{nerf2020eccv}
Mildenhall, B., Srinivasan, P.P., Tancik, M., Barron, J.T., Ramamoorthi, R., Ng, R.: Nerf: Representing scenes as neural radiance fields for view synthesis. Communications of the ACM  \textbf{65}(1),  99--106 (2021)

\bibitem{mle2003}
Myung, I.J.: Tutorial on maximum likelihood estimation. Journal of mathematical Psychology  \textbf{47}(1),  90--100 (2003)

\bibitem{gamma}
{National Institute of Standards and Technology}: Gamma function (2025), \url{https://dlmf.nist.gov/5}, nIST Digital Library of Mathematical Functions, Release 1.1.10

\bibitem{dinov2}
Oquab, M., Darcet, T., Moutakanni, T., Vo, H.V., Szafraniec, M., Khalidov, V., Fernandez, P., Haziza, D., Massa, F., El-Nouby, A., Howes, R., Huang, P.Y., Xu, H., Sharma, V., Li, S.W., Galuba, W., Rabbat, M., Assran, M., Ballas, N., Synnaeve, G., Misra, I., Jegou, H., Mairal, J., Labatut, P., Joulin, A., Bojanowski, P.: Dinov2: Learning robust visual features without supervision (2023)

\bibitem{splatssc2025}
Qian, R., Cao, H., Deng, T., Yuan, S., Xie, L.: Splatssc: Decoupled depth-guided gaussian splatting for semantic scene completion. In: Proceedings of the AAAI Conference on Artificial Intelligence. vol.~40, pp. 8520--8528 (2026)

\bibitem{nyueccv2012}
Silberman, N., Hoiem, D., Kohli, P., Fergus, R.: Indoor segmentation and support inference from rgbd images. In: European conference on computer vision. pp. 746--760. Springer (2012)

\bibitem{sscnet2017cvpr}
Song, S., Yu, F., Zeng, A., Chang, A.X., Savva, M., Funkhouser, T.: Semantic scene completion from a single depth image. In: Proceedings of the IEEE conference on computer vision and pattern recognition. pp. 1746--1754 (2017)

\bibitem{su2024roformer}
Su, J., Ahmed, M., Lu, Y., Pan, S., Bo, W., Liu, Y.: Roformer: Enhanced transformer with rotary position embedding. Neurocomputing  \textbf{568},  127063 (2024)

\bibitem{splatter2024cvpr}
Szymanowicz, S., Rupprecht, C., Vedaldi, A.: Splatter image: Ultra-fast single-view 3d reconstruction. In: Proceedings of the IEEE/CVF conference on computer vision and pattern recognition. pp. 10208--10217 (2024)

\bibitem{effnetv22021pmlr}
Tan, M., Le, Q.: Efficientnetv2: Smaller models and faster training. In: International conference on machine learning. pp. 10096--10106. PMLR (2021)

\bibitem{sparseocc2024cvpr}
Tang, P., Wang, Z., Wang, G., Zheng, J., Ren, X., Feng, B., Ma, C.: Sparseocc: Rethinking sparse latent representation for vision-based semantic occupancy prediction. In: Proceedings of the IEEE/CVF Conference on Computer Vision and Pattern Recognition (2024)

\bibitem{transformer}
Vaswani, A., Shazeer, N., Parmar, N., Uszkoreit, J., Jones, L., Gomez, A.N., Kaiser, {\L}., Polosukhin, I.: Attention is all you need. Advances in neural information processing systems  \textbf{30} (2017)

\bibitem{embodiedoccpp2025}
Wang, H., Wei, X., Zhang, X., Li, J., Bai, C., Li, Y., Lu, M., Zheng, W., Zhang, S.: Embodiedocc++: Boosting embodied 3d occupancy prediction with plane regularization and uncertainty sampler. In: Proceedings of the 33rd ACM International Conference on Multimedia. pp. 925--934 (2025)

\bibitem{opus2024nips}
Wang, J., Liu, Z., Meng, Q., Yan, L., Wang, K., Yang, J., Liu, W., Hou, Q., Cheng, M.: Opus: occupancy prediction using a sparse set. In: Advances in Neural Information Processing Systems (2024)

\bibitem{openocc2023iccv}
Wang, X., Zhu, Z., Xu, W., Zhang, Y., Wei, Y., Chi, X., Ye, Y., Du, D., Lu, J., Wang, X.: Openoccupancy: A large scale benchmark for surrounding semantic occupancy perception. In: Proceedings of the IEEE/CVF International Conference on Computer Vision. pp. 17850--17859 (2023)

\bibitem{bts2023cvpr}
Wimbauer, F., Yang, N., Rupprecht, C., Cremers, D.: Behind the scenes: Density fields for single view reconstruction. In: Proceedings of the IEEE/CVF Conference on Computer Vision and Pattern Recognition. pp. 9076--9086 (2023)

\bibitem{embodiedocc2025iccv}
Wu, Y., Zheng, W., Zuo, S., Huang, Y., Zhou, J., Lu, J.: Embodiedocc: Embodied 3d occupancy prediction for vision-based online scene understanding. In: Proceedings of the IEEE/CVF International Conference on Computer Vision (ICCV). pp. 26360--26370 (October 2025)

\bibitem{dav22024nips}
Yang, L., Kang, B., Huang, Z., Zhao, Z., Xu, X., Feng, J., Zhao, H.: Depth anything v2. Advances in Neural Information Processing Systems  \textbf{37},  21875--21911 (2024)

\bibitem{ndcscene2023iccv}
Yao, J., Li, C., Sun, K., Cai, Y., Li, H., Ouyang, W., Li, H.: Ndc-scene: Boost monocular 3d semantic scene completion in normalized device coordinates space. In: 2023 IEEE/CVF International Conference on Computer Vision (ICCV). pp. 9421--9431. IEEE Computer Society (2023)

\bibitem{iso2024eccv}
Yu, H., Wang, Y., Chen, Y., Zhang, Z.: Monocular occupancy prediction for scalable indoor scenes. In: Computer Vision – ECCV 2024: 18th European Conference, Milan, Italy, September 29–October 4, 2024, Proceedings, Part XXX. p. 38–54. Springer-Verlag, Berlin, Heidelberg (2024). \doi{10.1007/978-3-031-73404-5_3}, \url{https://doi.org/10.1007/978-3-031-73404-5_3}

\bibitem{flashocc2023}
Yu, Z., Shu, C., Deng, J., Lu, K., Liu, Z., Yu, J., Yang, D., Li, H., Chen, Y.: Flashocc: Fast and memory-efficient occupancy prediction via channel-to-height plugin  (2023)

\bibitem{voxelmamba2024nips}
Zhang, G., Fan, L., He, C., Lei, Z., Zhang, Z., Zhang, L.: Voxel mamba: Group-free state space models for point cloud based 3d object detection. Advances in Neural Information Processing Systems  \textbf{37},  81489--81509 (2024)

\bibitem{gslrm2024eccv}
Zhang, K., Bi, S., Tan, H., Xiangli, Y., Zhao, N., Sunkavalli, K., Xu, Z.: Gs-lrm: Large reconstruction model for 3d gaussian splatting. In: European Conference on Computer Vision. pp. 1--19. Springer (2024)

\bibitem{voxelsplat2025cvpr}
Zhu, Z., Wang, S., Xie, J., Liu, J.J., Wang, J., Yang, J.: Voxelsplat: Dynamic gaussian splatting as an effective loss for occupancy and flow prediction. In: Proceedings of the IEEE/CVF Conference on Computer Vision and Pattern Recognition (CVPR) (June 2025), to appear

\bibitem{quadricformer2025}
Zuo, S., Zheng, W., Han, X., Yang, L., Pan, Y., Lu, J.: Quadricformer: Scene as superquadrics for 3d semantic occupancy prediction. In: Advances in Neural Information Processing Systems (2025)

\bibitem{zuo2025gaussianworld}
Zuo, S., Zheng, W., Huang, Y., Zhou, J., Lu, J.: Gaussianworld: Gaussian world model for streaming 3d occupancy prediction. In: Proceedings of the Computer Vision and Pattern Recognition Conference. pp. 6772--6781 (2025)

\end{thebibliography}
